\documentclass[runningheads]{llncs}
\usepackage{graphicx}
\usepackage{xcolor}
\usepackage{array}
\usepackage{arydshln}
\usepackage{comment} 
\usepackage{tabularx}
\usepackage{subcaption}
\newcommand{\todo}[1]{
	\textcolor{red}{[\textbf{ToDo}: \emph{#1}]}
}
\newcommand{\todiscuss}[1]{
	\textcolor{orange}{[\textbf{ToDiscuss}: \emph{#1}]}
}

\usepackage{varioref}
\usepackage{prettyref}
\usepackage[strings]{underscore}
\usepackage[numbers]{natbib}
\usepackage{float}
\usepackage{mathtools}
\usepackage{enumitem}
\usepackage{multirow}
\usepackage{amsmath}
\usepackage{amssymb}
\usepackage{graphicx}

\usepackage{mathalfa}
\usepackage{lscape}
\usepackage{relsize}

\usepackage{listings} %R-Code einbinden.
\usepackage{booktabs} % Bessere Tabellen.
\usepackage{siunitx}
\usepackage{algorithm}
\usepackage{algpseudocode}
\usepackage{authblk} % for author affiliation

\global\long\def\phantomeq{\mathrel{\phantom{=}}}%

\global\long\def\nodeSetNull{\overline{J}_{0}}%
\global\long\def\nodes{J}%
\global\long\def\nodeSet{\overline{J}}%

\global\long\def\robotnumb{N}%

\global\long\def\poolSize{N}%

\global\long\def\routingMatrix{\mathcal{R}}%
\global\long\def\routingProb{r}%

\global\long\def\nodeProc{Y}%
\global\long\def\nodeProcALL{\mathbf{Y}}%

\global\long\def\nodeQueue{n}%
\global\long\def\nodeServiceRate{\nu}%

\global\long\def\externalQueueProc{X_{\text{ex}}}%
\global\long\def\externalQueue{n_{\text{ex}}}%

\global\long\def\arrival{\lambda_{\text{BO}}}%
\global\long\def\arrivalLS{\lambda_{\text{LC}}}%
\global\long\def\customerOrdersArrival{\lambda_{\text{CO}}}%

\global\long\def\procbackordering{X_{\text{ex}}}%
\global\long\def\backordering{n_{\text{ex}}}%

\global\long\def\procfreerobi{Y_{\text{idle robots}}}%
\global\long\def\freerobi{k}%

\global\long\def\procToPod{Y_{\text{to pod}}}%
\global\long\def\toPod{m_{\text{1}}}%
\global\long\def\transToPod{\nu_{1}}%
\global\long\def\phiToPod{\phi_{1}}%

\global\long\def\procfreerobi{Y_{0}}%
\global\long\def\freerobi{n_{0}}%
\global\long\def\freerobiState{0}%

\global\long\def\procToPod{Y_{\text{sp}}}%
\global\long\def\toPod{n_{\text{sp}}}%
\global\long\def\toPodState{\text{sp}}%
\global\long\def\transToPod{\mu_{\text{sp}}}%
\global\long\def\phiToPod{\phi_{\text{sp}}}%

\global\long\def\procToPickOne{Y_{\text{pp}_{1}}}%
\global\long\def\toPickOneState{\text{pp}_{1}}%
\global\long\def\toPickOne{n_{\text{pp}_{1}}}%
\global\long\def\transToPickerOne{\mu_{\text{pp}_{1}}}%
\global\long\def\phiToPickOne{\phi_{\text{pp}_{1}}}%
\global\long\def\transprobPickOne{q_{\text{pp}_{1}}}%

\global\long\def\procToPickTwo{Y_{\text{pp}_{2}}}%
\global\long\def\toPickTwoState{\text{pp}_{2}}%
\global\long\def\toPickTwo{n_{\text{pp}_{2}}}%
\global\long\def\transToPickerTwo{\mu_{\text{pp}_{2}}}%
\global\long\def\phiToPickTwo{\phi_{\text{pp}_{2}}}%
\global\long\def\transprobPickTwo{q_{\text{pp}_{2}}}%

\global\long\def\procPickerQueueOne{Y_{\text{p}_{1}}}%
\global\long\def\pickerQueueOne{n_{\text{p}_{1}}}%
\global\long\def\pickerQueueOneState{\text{p}_{1}}%
\global\long\def\pickingOne{\nu_{\text{p}_{1}}}%
\global\long\def\transprobReplOne{q_{\text{p}_{1}\text{r}_1}}%
\global\long\def\transprobpickReplOne{q_{\text{p}_{1}\text{s}}}%

\global\long\def\procPickerQueueTwo{Y_{\text{p}_{2}}}%
\global\long\def\pickerQueueTwoState{\text{p}_{2}}%
\global\long\def\pickerQueueTwo{n_{\text{p}_{2}}}%
\global\long\def\pickingTwo{\nu_{\text{p}_{2}}}%
\global\long\def\transprobReplTwo{q_{\text{p}_{2}\text{r}_{2}}}%
\global\long\def\transprobpickReplTwo{q_{\text{p}_{2}\text{s}}}%

\global\long\def\procToStorageOne{Y_{\text{p}_{1}\text{s}}}%
\global\long\def\toStorageOneState{\text{p}_{1}\text{s}}%
\global\long\def\toStorageOne{n_{\text{p}_{1}\text{s}}}%
\global\long\def\transToStorageOne{\mu_{\text{p}_{1}\text{s}}}%
\global\long\def\phiToStorageOne{\phi_{\text{p}_{1}\text{s}}}%

\global\long\def\procToStorageTwo{Y_{\text{p}_{2}\text{s}}}%
\global\long\def\toStorageTwoState{\text{p}_{2}\text{s}}%
\global\long\def\toStorageTwo{n_{\text{p}_{2}\text{s}}}%
\global\long\def\transToStorageTwo{\mu_{\text{p}_{2}\text{s}}}%
\global\long\def\phiToStorageTwo{\phi_{\text{p}_{2}\text{s}}}%

\global\long\def\toReplOne{n_{\text{p}_{1}\text{r}}}%old
\global\long\def\procToReplOneNew{Y_{\text{p}_{1}\text{r}_{1}}}%new
\global\long\def\toReplOneStateNew{\text{p}_{1}\text{r}_{1}}%new
\global\long\def\toReplOneNew{n_{\text{p}_{1}\text{r}_{1}}}%new
\global\long\def\transToReplOneNew{\mu_{\text{p}_{1}\text{r}_{1}}}%new
\global\long\def\phiToReplOneNew{\phi_{\text{p}_{1}\text{r}_{1}}}%new

\global\long\def\procToReplTwoNew{Y_{\text{p}_{2}\text{r}_{2}}}%new
\global\long\def\toReplTwoStateNew{\text{p}_{2}\text{r}_{2}}%new
\global\long\def\toReplTwoNew{n_{\text{p}_{2}\text{r}_{2}}}%new
\global\long\def\transToReplTwoNew{\mu_{\text{p}_{2}\text{r}_{2}}}%new
\global\long\def\phiToReplTwoNew{\phi_{\text{p}_{2}\text{r}_{2}}}%new

\global\long\def\procReplOne{Y_{\text{r}_{1}}}%new
\global\long\def\replQueueOneState{\text{r}_{1}}%new
\global\long\def\replQueueOne{n_{\text{r}_{1}}}%new
\global\long\def\replOne{\nu_{r_{1}}}%new

\global\long\def\procReplTwo{Y_{\text{r}_{2}}}%new
\global\long\def\replQueueTwoState{\text{r}_{2}}%new
\global\long\def\replQueueTwo{n_{\text{r}_{2}}}%new
\global\long\def\replTwo{\nu_{r_{2}}}%new

\global\long\def\toReplToStorageState{\text{rs}}%old
\global\long\def\transReplToStorage{\mu_{\text{rs}}}%old
\global\long\def\procReplOneToStorage{Y_{\text{r}_{1}\text{s}}}%new
\global\long\def\toReplOneToStorageState{\text{r}_{1}\text{s}}%new
\global\long\def\toReplOneToStorage{n_{\text{r}_{1}\text{s}}}%new
\global\long\def\transReplOneToStorage{\mu_{\text{r}_{1}\text{s}}}%new
\global\long\def\phiToStorageOne{\phi_{\text{r}_{1}\text{s}}}%new

\global\long\def\procReplTwoToStorage{Y_{\text{r}_{2}\text{s}}}%new
\global\long\def\toReplTwoToStorageState{\text{r}_{2}\text{s}}%new
\global\long\def\toReplTwoToStorage{n_{\text{r}_{2}\text{s}}}%new
\global\long\def\transReplTwoToStorage{\mu_{\text{r}_{2}\text{s}}}%new
\global\long\def\phiToStorageTwo{\phi_{\text{r}_{2}\text{s}}}%new

\global\long\def\freerobi{k_{\text{idle robots}}}%

\global\long\def\podToOrderRatio{\sigma_{\text{pod/order}}}%

\global\long\def\TOtask{TO_{\text{task}}}%
\begin{document}
\title{Introducing Combi-Stations in Robotic Mobile Fulfilment Systems: A Queueing-Theory-Based Efficiency Analysis}
\titlerunning{Introducing Combi-Stations in Robotic Mobile Fulfilment Systems}
% If the paper title is too long for the running head, you can set
% an abbreviated paper title here
%
\author{Lin Xie\inst{1}\orcidID{0000-0002-3168-4922} \and
Sonja Otten\inst{1}\orcidID{0000-0002-3124-832X}}
\authorrunning{Xie and Otten}
% First names are abbreviated in the running head.
% If there are more than two authors, 'et al.' is used.
%
\institute{University of Twente,  Drienerlolaan 5, 7522 NB Enschede, The Netherlands \\
\email{\{lin.xie,s.otten\}@utwente.nl}}
\maketitle              % typeset the header of the contribution
\begin{abstract}
%The abstract should briefly summarize the contents of the paper in 15--250 words.
In the era of digital commerce, the surge in online shopping and the expectation for rapid delivery have placed unprecedented demands on warehouse operations. The traditional method of order fulfilment, where human order pickers traverse large storage areas to pick items, has become a bottleneck, consuming valuable time and resources. Robotic Mobile Fulfilment Systems (RMFS) offer a solution by using robots to transport storage racks directly to human-operated picking stations, eliminating the need for pickers to travel. This paper introduces `combi-stations'—a novel type of station that enables both item picking and replenishment, as opposed to traditional separate stations. We analyse the efficiency of combi-stations using queueing theory and demonstrate their potential to streamline warehouse operations. Our results suggest that combi-stations can reduce the number of robots required for stability and significantly reduce order turnover time, indicating a promising direction for future warehouse automation.
\end{abstract}
%
%Full papers should have a length between 8 to 15 pages following the LNCS style files. For a paper to be included in this volume, it should be presented at the conference by one of the authors. Furthermore, the authors are required to use the submission form available here to:

%Upload the pdf file of the 8-15 page regular paper in LNCS style on the Easychair platform.
%Paste the abstract corresponding with the full paper in the textbox (maximum 3000 characters, including spaces) to be included in the abstracts booklet, which will be distributed during the conference.

%
\section{Introduction}
According to a recent report by Statista (2024)\footnote{https://www.statista.com/statistics/379046/worldwide-retail-e-commerce-sales/}, global retail e-commerce sales reached \$5.8 trillion in 2023 and are expected to exceed \$8 trillion by 2027. In today's fast-paced economy, timely order fulfilment is critical. To accommodate this rapid growth, warehouses must operate more efficiently by turning pallets into ready-to-ship packages.
The primary and most time-consuming task in a warehouse is to pick items from their storage locations to fulfil customer orders (called order picking). This process can account for around 50-65\% of operating costs. Improving the efficiency of this process is therefore paramount (see \cite{de2007design}).  In a traditional manual order picking system, also known as a picker-to-parts system, pickers spend about 70\% of their working time on unproductive tasks such as searching and travelling (see \cite{Tompkins.2010}). To minimise the travelling time of human pickers, many solutions have been proposed in the literature and in practice, such as zoning \cite{de2012determining}, mixed-shevles storage \cite{weidinger2018storage}, splitting orders during picking \cite{xie2019split}, using co-robots to do the travelling \cite{xie2020rafs} and various automated systems (see \cite{boysen2019warehousing} for an overview).

\begin{figure}[H]
    \begin{subfigure}{\textwidth}
        \centering
        \includegraphics[width=\linewidth]{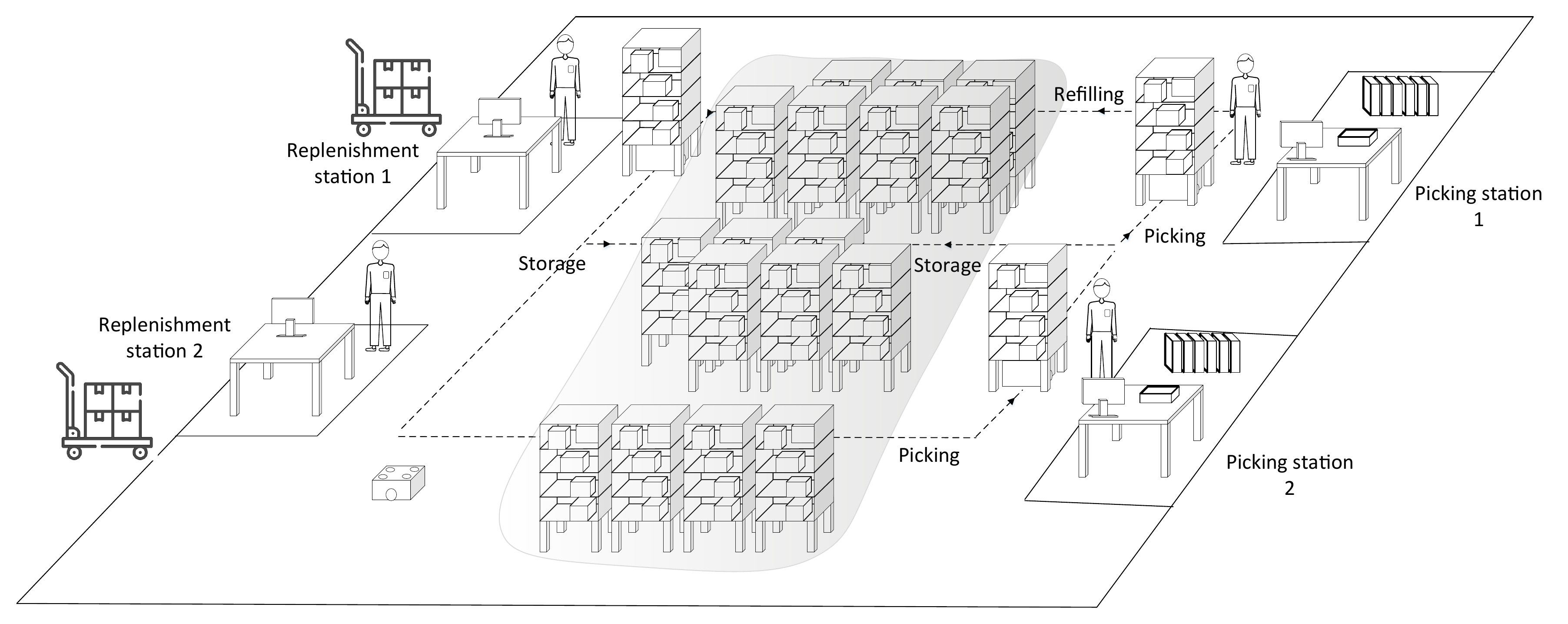}  % Replace 'image1' with your actual image filename
        \caption{The picking and replenishment processes of two-station types.}
        \label{fig:two-station-types}
    \end{subfigure}%
    \hfill
    \begin{subfigure}{\textwidth}
        \centering
        \includegraphics[width=\linewidth]{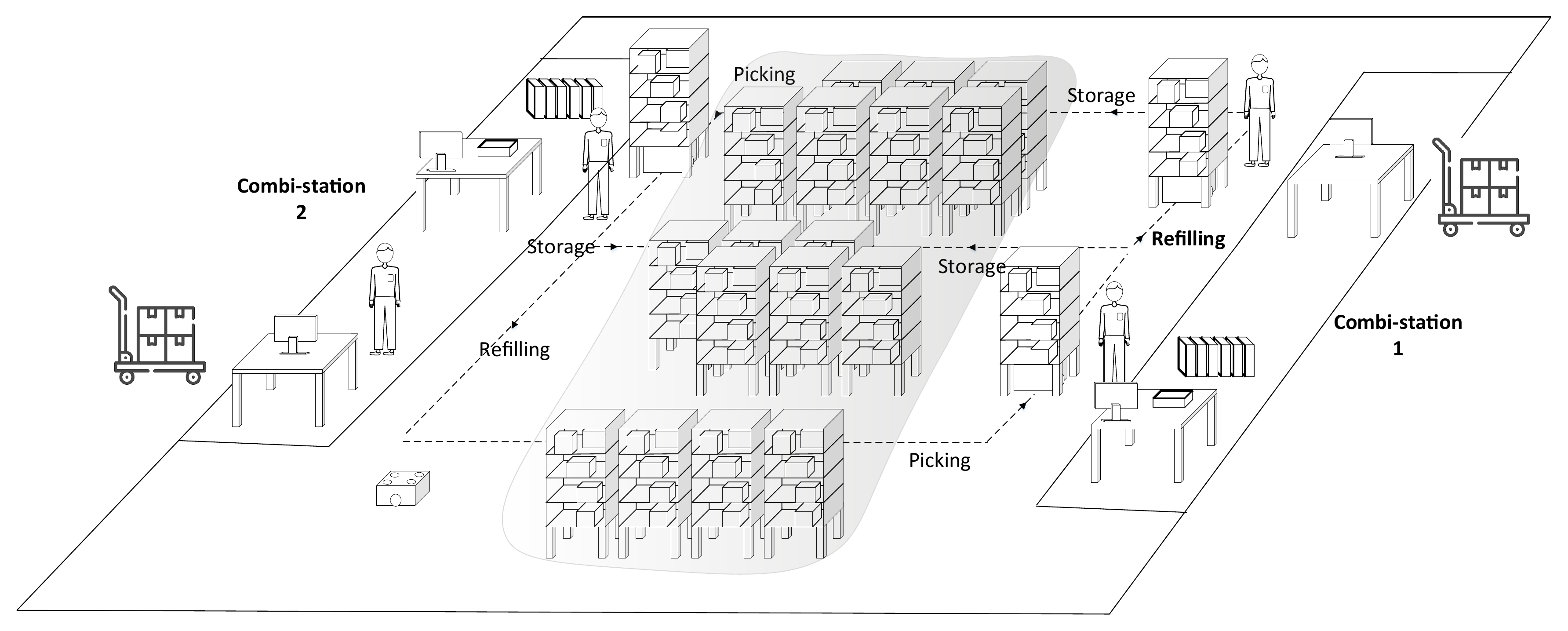}  % Replace 'image2' with your actual image filename
        \caption{The picking and replenishment processes of combi-stations.}
        \label{fig:combi-station}
    \end{subfigure}
    \caption{The picking and replenishment processes of two RMFS systems.}
    \label{fig:rmfs}
\end{figure}
In this paper, we consider one such automated system, the Robotic Mobile Fulfilment System (RMFS), which was developed by Kiva Systems LLC, now Amazon Robotics LLC. In such a system (a small example is depicted in Figure~\ref{fig:two-station-types}), robots are sent to retrieve pods (also called racks or shelves) from the storage area (shown in the grey area of Figure~\ref{fig:two-station-types}) and bring them to human operators at picking stations (located on the right side of Figure~\ref{fig:two-station-types}), where the items are picked according to customer orders. 
After picking, the robots return the pods to the storage area or to replenishment stations (located on the left side of Figure~\ref{fig:two-station-types}) for replenishment before returning to the storage area. Sometimes packing is also done at the picking stations. More often the picked items are transported (e.g.~by conveyors or mobile robots) to packing stations for final packing. However, packing stations are not considered in this paper. 
In Figure~\ref{fig:combi-station}, there is a combi-station to the right and left of the storage area. Each combi-station is designed with dual functionality, featuring distinct sections for both picking and replenishment tasks. This layout allows for immediate replenishment following the picking process within the same station, streamlining operations by eliminating the need to traverse to a different location within the storage area. Immediate replenishment is optional and does not need to be performed after every picking activity. In both cases in Figure~\ref{fig:rmfs}, a robot has to wait if the human operator at its destination is busy (i.e.~a queue is formed). For a fair comparison, we assume that in both cases in Figure~\ref{fig:rmfs} we have two picking and two replenishment human operators. 

There are some publications related to the layout design in the RMFS, such as the dimension of the storage area \cite{lamballais2019inventory}, the shape of the forward area \cite{aldarondo2022expected}, the number of pods, the ratio of stations, the placement of stations \cite{lamballais2017estimating, lamballais2019inventory} and the number of robots \cite{gong2021robotic, otten2019numRobot, zou2018battery}. The most common types of stations are picking stations, where human pickers pick items from pods, and replenishment stations, where items are stored on pods. They are usually separated so that the replenishment stations are close to the inbound (pallet) door, while the picking stations are closer to the outbound (parcel) door. However, automation can make their placement more flexible. For example, mobile robots (MIR robots, fetch robots, Amazon Proteus, etc.) are used to transport parcels or pallets within the warehouse.

In this paper, we present a new type of station, called a combi-station, which allows both picking and replenishment. Furthermore, to evaluate the efficiency of using combi-stations, we model the RMFS as a semi-open queueing network with backordering (SOQN-BO) and apply the approximation methods proposed in \cite{otten2019numRobot} to evaluate its efficiency. More specifically, we compute the average turnover time, i.e.~the time from the arrival of an order at the system to the completion of picking.

In the following,  we model the RMFS with two-station types and the RMFS with combi-stations as SOQN-BO in Section~\ref{sec:model} and include the approximation methods from \cite{otten2019numRobot} to calculate the  turnover time. In Section~\ref{sec:results}, we present some computational results  related to our investigation. Finally, our paper concludes with a short summary in Section~\ref{sec:concl}.

 \section{Modelling as an SOQN-BO} \label{sec:model}
In Subsection~\ref{subsec:fromoldpaper},  we first give a brief description of an SOQN-BO as outlined in \cite{otten2019numRobot}. Then we model the example depicted in Figure~\ref{fig:two-station-types} (an RMFS with two-station types) as an SOQN-BO in Subsection~\ref{subsec:rmfs_modell_two_types}. Based on this model, we also model the example shown in Figure~\ref{fig:combi-station} (an RMFS with combi-stations) in Subsection~\ref{subsec:model_combi}.  Finally, we briefly introduce the approximation methods from \cite{otten2019numRobot} that we employ for calculating the turnover time in Subsection~\ref{subsec:turnover}.
%The approximation methods we use from \cite{otten2019numRobot} to calculate the turnover time are shown in Subsection~\ref{subsec:turnover}.

%---------
\subsection{Description of an SOQN-BO} \label{subsec:fromoldpaper}

\begin{figure}[h]
	\center
	\includegraphics[width=0.85\textwidth]{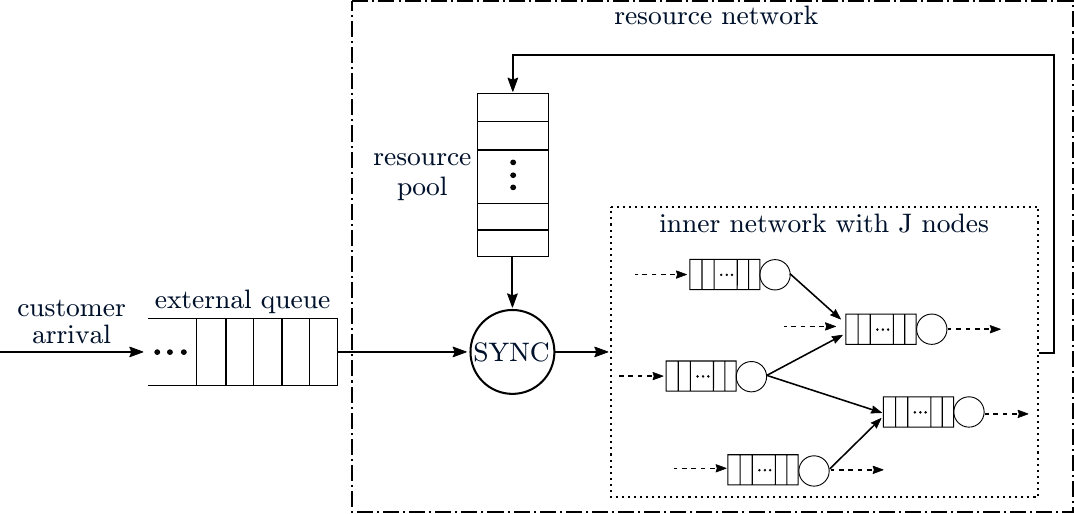}
	\caption{An SOQN with backordering and external queue \cite[Fig.~2]{otten2019numRobot}. }\label{fig:SOQN}
\end{figure}

A semi-open queueing network has characteristics of both an open queueing network and a closed queueing network.
Figure~\ref{fig:SOQN} shows an SOQN-BO as described in \cite[Section 2.1]{otten2019numRobot}. It consists of a queueing network (``inner network''), a resource pool, and an external queue.

In this system, customers arrive following a Poisson process with a rate $\arrival>0$. For service each customer requires exactly one resource from the resource pool. If a resource is available when a customer arrives, the resource enters the inner network to complete the customer's order. However, if a resource is not available when a customer arrives, the new customer waits in the external queue on a first-come, first-served (FCFS) basis until a resource becomes available (backordering).

When the resource leaves the inner network, it returns to the resource pool (referred to as node $0$) and waits for the next customer to arrive. Whenever the external queue is not empty and a resource item is
returned to the resource pool, it is immediately synchronised with the customer at the front of the queue. Therefore, the movement of resources within this system forms a closed network, aptly named the resource network. The maximum number of resources available in the pool is denoted by $\poolSize$.

The inner network comprises $\nodes\geq1$ numbered service stations
(nodes), denoted by $\nodeSet:=\left\{ 1,\ldots,\nodes\right\} $.
Each station $j$ consists of a single server with infinite waiting
room under  FCFS regime or  processor sharing regime. Customers
in the network are indistinguishable. The service times follow an exponentially distributed random variable with mean  $1$. If there are $n_{j}>0$
customers present at node $j$, the service at node $j$ is provided with
intensity $\nodeServiceRate_{j}(\nodeQueue_{j})>0$. All service and
inter-arrival times constitute an independent family of random variables.

Movements of resources in the inner network follow a Markovian
routing mechanism: After synchronisation with a customer, a resource
visits node $j$ with probability $\routingProb(0,j)\geq0$. When leaving node $i$, a resource selects with probability $\routingProb(i,j)\geq0$
to visit node $j$ next. It then immediately enters node  $j$. 
If the server at node $j$ is idle, the resource starts its service. Otherwise, it joins the tail of the queue at node $j$.
This resource may also leave the inner
network with probability $\routingProb(i,0)\geq0$. It holds $\sum_{j=0}^{\nodes}\routingProb(i,j)=1$
with $\routingProb(0,0):=0$ for all $i\in\nodeSetNull:=\left\{ 0,1,\ldots,\nodes\right\} $.
The resource’s routing decision, given the departure node $i$, is independent of the network’s history.
 We assume that the routing matrix $\routingMatrix:=\left(\routingProb(i,j):i,j\in\nodeSetNull\right)$
is irreducible.

To obtain a Markovian process description, we introduce the following notation. We denote by $\externalQueueProc(t)$
the number of customers in the external queue at time $t\geq0$, by
$\nodeProc_{0}(t)$ the number of resources in the resource pool at
time $t\geq0$ and by $\nodeProc_{j}(t)$, $j\in\nodeSet$, the number
of resources present at node $j$ in the inner network at time $t\geq0$,
either waiting or in service. We call this $\nodeProc_{j}(t)$ queue
length at node $j\in\nodeSetNull$ at time $t\geq0$. Then $\nodeProcALL(t):=\left(\nodeProc_{j}(t):j\in\nodeSetNull\right)$
is the queue length vector of the resource network at time $t\geq0$.
We define the joint queue length process of the semi-open network
with \textbf{b}ack\textbf{o}rdering by
%\[ 
$
Z_{\text{BO}}:=\left(\left(\externalQueueProc(t),\nodeProcALL(t)\right):t\geq0\right).
$
%\]
Due to the assumptions of independence and memorylessness,\label{independence-memorylessness}
$Z_{\text{BO}}$ is an  irreducible Markov process with state space
\begin{align*}
E & :=\big\{\left(0,\nodeQueue_{0},\nodeQueue_{1},\ldots,\nodeQueue_{J}\right):\nodeQueue_{j}\in\left\{ 0,\ldots,\poolSize\right\} \:\forall j\in\nodeSetNull,\sum_{j\in\nodeSetNull}\nodeQueue_{j}=\poolSize\big\}\\
& \phantomeq\cup\big\{\left(\externalQueue,0,\nodeQueue_{1},\ldots,\nodeQueue_{J}\right):\externalQueue\in\mathbb{N},\:\nodeQueue_{j}\in\left\{ 0,\ldots,\poolSize\right\} \:\forall j\in\nodeSet,\sum_{j\in\nodeSet}\nodeQueue_{j}=\poolSize\big\}.
\end{align*}

%---------

 \subsection{Modelling the RMFS with two-station types} \label{subsec:rmfs_modell_two_types}

 To model the RMFS with two-station types as an SOQN-BO we use the same definition of robot tasks as in \cite[Section 5.2]{otten2019numRobot}. A robot's task is not exactly the same as a customer's order, because the items in the order may be spread across several pods. Some orders can share the same pod. The customer orders are split into items as introduced by \cite{xie2019split}, so the robot's tasks are a stream of ``bring a pod to a picking station''. The task stream is modelled as a Poisson stream with rate $\arrival=\customerOrdersArrival\cdot\podToOrderRatio$, where the order arrival rate $\customerOrdersArrival$ and the average pod/order ratio $\podToOrderRatio$ are given.
 
 Since RMFS is open with respect to tasks and closed with respect to robots, which are the resources, we can model the example shown in Figure~\ref{fig:two-station-types} as SOQN-BO in the following way.
\begin{figure}[h]
	\includegraphics[width=1\textwidth]{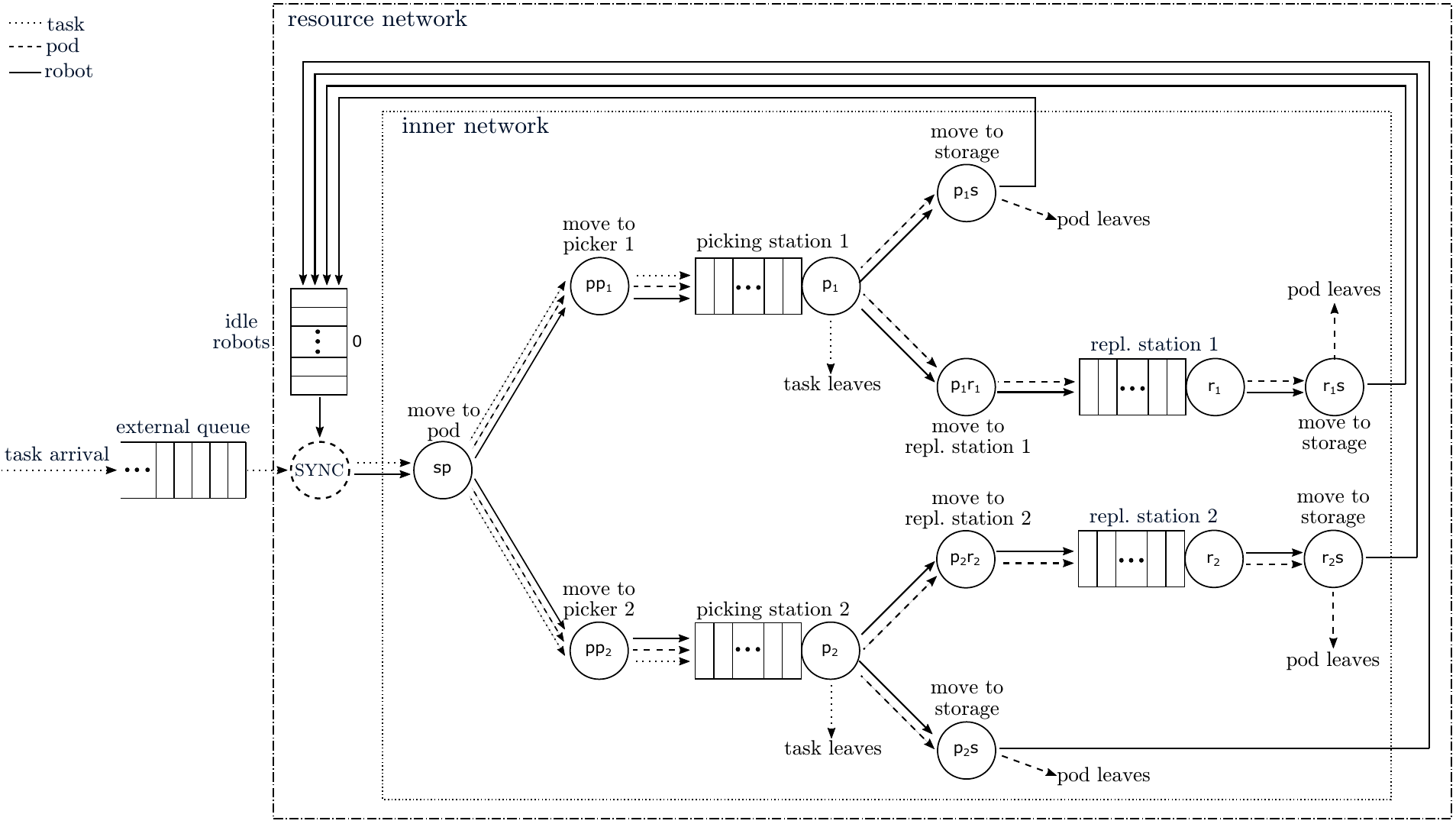}
	\caption{RMFS with two-station types modelled as an SOQN-BO.}
	\label{fig:model-two-station-types}
\end{figure}

As shown in Figure~\ref{fig:model-two-station-types}, each task requires exactly one idle robot from the robot pool (resource pool) to enter the inner network,
which is referred to as node $0$. 
If no idle robot is available, the new task must wait in an external queue until a robot becomes available ("backordering"). The maximum number of robots in the resource
pool is $\robotnumb$. The inner network in the example
consists of 13 nodes, denoted by
\[
\nodeSet:=\left\{ \toPodState,\toPickOneState,\toPickTwoState,\pickerQueueOneState,\pickerQueueTwoState,\toStorageOneState,\toStorageTwoState,\toReplOneStateNew,\toReplTwoStateNew,\replQueueOneState, \replQueueTwoState,\toReplOneToStorageState,\toReplTwoToStorageState\right\} .
\]
The meaning and notations of nodes are given in \prettyref{tab:model-nodes-table}.
The robot with the assigned task moves through the network.

From the perspective of a robot, the following processes take place as it moves through the network:

\begin{itemize}
	\item  The idle robot awaits assignment to a  task (bring a particular pod).
	\item The robot moves with the assigned task to a pod.
	\item With this pod the robot moves  with probability $\transprobPickOne\in\left(0,1\right)$
	to picking station $1$ and with probability $\transprobPickTwo\in\left(0,1\right)$
	to station $2$, $\transprobPickOne+\transprobPickTwo=1$.
	\item At the picking stations, the robot queues with the pod.
	\item After picking at picking station $1$ resp.~picking station 2, the robot faces two possibilities: \\
 Option A:
		\begin{itemize}
			\item[-] The robot carries the pod directly back to the storage area with probability
			$\transprobpickReplOne\in\left(0,1\right)$ resp.~$\transprobpickReplTwo\in\left(0,1\right)$ and
            \item[-] waits for the next task.
        \end{itemize}
    \item[$\ $] Option B:
        \begin{itemize}
			\item[-] The robot moves to the replenishment station with probability $\transprobReplOne\in\left(0,1\right)$
			resp.~$\transprobReplTwo\in\left(0,1\right)$, whereby $\transprobpickReplOne+\transprobReplOne=1$
			resp.~$\transprobpickReplTwo+\transprobReplTwo=1$, 
		  \item[-] queues at the replenishment station, and
		  \item[-] carries the pod back to the storage area and 
            \item[-] waits for the next task.
        \end{itemize}
\end{itemize}

Each of these processes is modelled as a queue.
The movements of the robots are modelled by processor-sharing
nodes with exponential service times with intensities
$\nu_{j}(n_{j}):=\mu_{j}\cdot\phi_{j}(n_{j})$, $j\in\nodeSet\setminus\left\{ \pickerQueueOneState,\pickerQueueTwoState,\replQueueOneState, \replQueueTwoState\right\} $,
presented in Table \ref{tab:model-nodes-table}.

The two picking stations and two replenishment stations, which are
referred to as nodes $\pickerQueueOneState$ and $\pickerQueueTwoState$
resp.~nodes $\replQueueOneState$ and $\replQueueTwoState$, consist of a single server with waiting room under the FCFS regime. The picking times and the replenishment
times are exponentially distributed with rates $\pickingOne$ and $\pickingTwo$
resp.~$\replOne$ and $\replTwo$.

The robots travel among the nodes following an irreducible routing matrix
$\routingMatrix:=\left(\routingProb(i,j):i,j\in\nodeSetNull\right)$,
whereby $\nodeSetNull:=\left\{ 0\right\} \cup\nodeSet$, which is
given by
\[
\routingMatrix=\left(\begin{array}{c|cccccccccccccc}
& \freerobiState & \toPodState & \toPickOneState & \toPickTwoState & \pickerQueueOneState & \pickerQueueTwoState & \toStorageOneState & \toStorageTwoState & \toReplOneStateNew & \toReplTwoStateNew & \replQueueOneState & \replQueueTwoState & \toReplOneToStorageState & \toReplTwoToStorageState\\
\hline \freerobiState &  & 1\\
\toPodState &  &  & \transprobPickOne & \transprobPickTwo\\
\toPickOneState &  &  &  &  & 1\\
\toPickTwoState &  &  &  &  &  & 1\\
\pickerQueueOneState &  &  &  &  &  &  & \transprobpickReplOne &  & \transprobReplOne\\
\pickerQueueTwoState &  &  &  &  &  &  &  & \transprobpickReplTwo &  & \transprobReplTwo\\
\toStorageOneState & 1\\
\toStorageTwoState & 1\\
\toReplOneStateNew &  &  &  &  &  &  &  &  &  &  & 1\\
\toReplTwoStateNew &  &  &  &  &  &  &  &  &  &  & & 1\\
\replQueueOneState &  &  &  &  &  &  &  &  &  &  &  & & 1\\
\replQueueTwoState &  &  &  &  &  &  &  &  &  &  &  & & & 1\\
\toReplOneToStorageState & 1\\
\toReplTwoToStorageState & 1
\end{array}\right).
\]

We define the joint stochastic process $Z$ of this system by
\begin{align*}
Z & :=\Big(\Big(\procbackordering(t),\procfreerobi(t),\procToPod(t),\procToPickOne(t),\procToPickTwo(t),\procPickerQueueOne(t),\procPickerQueueTwo(t),\procToStorageOne(t),\procToStorageTwo(t),\\
& \phantomeq\quad\ \ \procToReplOneNew(t),\procToReplTwoNew(t),\procReplOne(t),\procReplTwo(t),\procReplOneToStorage(t),\procReplTwoToStorage(t)\Big):t\geq0\Big).
\end{align*}

Due to the usual independence and memoryless assumptions, 
%(see
%the assumptions \vpageref{independence-memorylessness}), 
$Z$ is
an irreducible Markov process with state space
\begin{align*}
E & :=\phantom{\cup}\big\{\left(0,\freerobi,\toPod,\toPickOne,\toPickTwo,\pickerQueueOne,\pickerQueueTwo,\toStorageOne,\toStorageTwo,\toReplOneNew,\toReplTwoNew,\replQueueOne, \replQueueTwo,\toReplOneToStorage, \toReplTwoToStorage\right):\\
& \phantomeq\phantom{\cup\big\{\:}n_{j}\in\left\{ 0,\ldots,\robotnumb\right\} \ \forall j\in\nodeSetNull,\ \sum_{j\in\nodeSetNull}n_{j}=\robotnumb\big\}\\
& \phantomeq\cup\big\{\left(\backordering,0,\toPod,\toPickOne,\toPickTwo,\pickerQueueOne,\pickerQueueTwo,\toStorageOne,\toStorageTwo,\toReplOneNew,\toReplTwoNew,\replQueueOne, \replQueueTwo,\toReplOneToStorage, \toReplTwoToStorage\right):\\
& \phantomeq\phantom{\cup\big\{\:}\backordering\in\mathbb{N},\:n_{j}\in\left\{ 0,\ldots,\robotnumb\right\} \ \forall j\in\nodeSet,\ \sum_{j\in\nodeSet}n_{j}=\robotnumb\big\}.
\end{align*}

\subsection{Modelling the RMFS with combi-stations} \label{subsec:model_combi}
We adapt the model in Figure~\ref{fig:model-two-station-types} to the case with two combi-stations. The main difference is that there are no nodes representing movement to the replenishment stations $\toReplOneStateNew$ and $\toReplTwoStateNew$. It can also be interpreted that the average travel time of $\toReplOneStateNew$ and $\toReplTwoStateNew$ is equal to zero. However, since this is not allowed, the model with combi-stations cannot be considered as a special case of the model with two-station types. Nevertheless, it can be considered as a limit case when the average travel time approaches zero.

\begin{figure}[h]
	\includegraphics[width=1\textwidth]{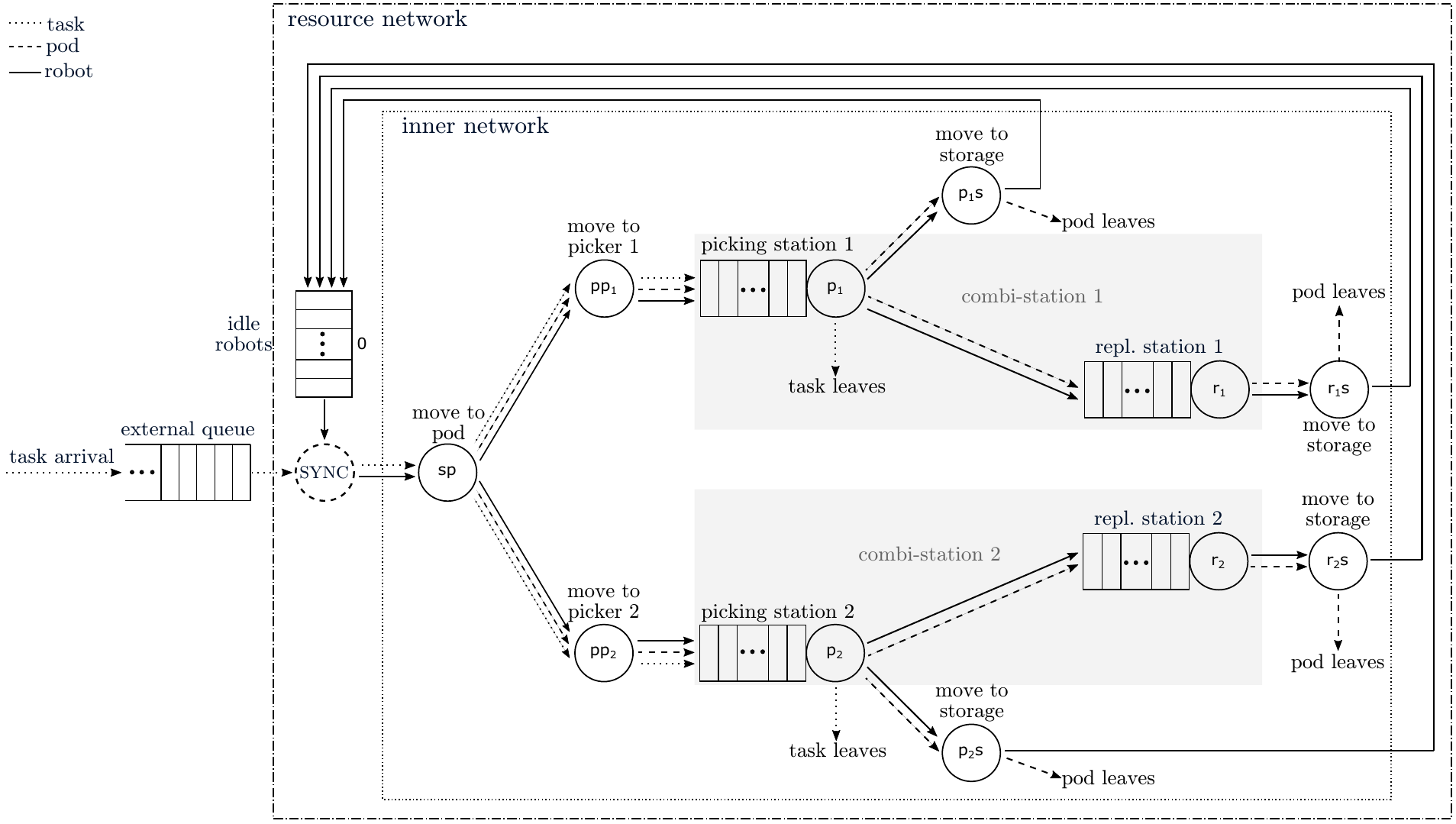}
	\caption{RMFS with combi-stations modelled as an SOQN-BO.}
	\label{fig:model_combi}
\end{figure}

As shown in Figure~\ref{fig:model_combi}, we retain the picking and replenishment station nodes, but for the sake of clarity these are the picking and replenishment parts within combi-stations 1 and 2. 

The definition of the nodes is the same as described in Table~\ref{tab:model-nodes-table}. The inner network in the example
has two nodes less than the model with two types of stations, namely 11 nodes, denoted by

\[
\nodeSet:=\left\{ \toPodState,\toPickOneState,\toPickTwoState,\pickerQueueOneState,\pickerQueueTwoState,\toStorageOneState,\toStorageTwoState,\replQueueOneState, \replQueueTwoState,\toReplOneToStorageState,\toReplTwoToStorageState\right\} .
\]

We update the routing matrix $\routingMatrix$ and the joint stochastic process $Z$ correspondingly without $\toReplOneStateNew$ and $\toReplTwoStateNew$ (nodes representing movement to the
replenishment stations), namely
\[
\routingMatrix=\left(\begin{array}{c|cccccccccccc}
& \freerobiState & \toPodState & \toPickOneState & \toPickTwoState & \pickerQueueOneState & \pickerQueueTwoState & \toStorageOneState & \toStorageTwoState &  \replQueueOneState & \replQueueTwoState & \toReplOneToStorageState & \toReplTwoToStorageState\\
\hline \freerobiState &  & 1\\
\toPodState &  &  & \transprobPickOne & \transprobPickTwo\\
\toPickOneState &  &  &  &  & 1\\
\toPickTwoState &  &  &  &  &  & 1\\
\pickerQueueOneState &  &  &  &  &  &  & \transprobpickReplOne &  & \transprobReplOne\\
\pickerQueueTwoState &  &  &  &  &  &  &  & \transprobpickReplTwo &  & \transprobReplTwo\\
\toStorageOneState & 1\\
\toStorageTwoState & 1\\
\replQueueOneState &  &  &  &  &  &  &  &  &  &  & 1 & \\
\replQueueTwoState &  &  &  &  &  &  &  &  &  &  &  &  1\\
\toReplOneToStorageState & 1\\
\toReplTwoToStorageState & 1
\end{array}\right)
\]

and 
\begin{align*}
Z & :=\Big(\Big(\procbackordering(t),\procfreerobi(t),\procToPod(t),\procToPickOne(t),\procToPickTwo(t),\procPickerQueueOne(t),\procPickerQueueTwo(t),\procToStorageOne(t),\procToStorageTwo(t),\\
& \phantomeq\quad\ \ \procReplOne(t),\procReplTwo(t),\procReplOneToStorage(t),\procReplTwoToStorage(t)\Big):t\geq0\Big)
\end{align*}
with state space
\begin{align*}
E & :=\phantom{\cup}\big\{\left(0,\freerobi,\toPod,\toPickOne,\toPickTwo,\pickerQueueOne,\pickerQueueTwo,\toStorageOne,\toStorageTwo,\replQueueOne, \replQueueTwo,\toReplOneToStorage, \toReplTwoToStorage\right):\\
& \phantomeq\phantom{\cup\big\{\:}n_{j}\in\left\{ 0,\ldots,\robotnumb\right\} \ \forall j\in\nodeSetNull,\ \sum_{j\in\nodeSetNull}n_{j}=\robotnumb\big\}\\
& \phantomeq\cup\big\{\left(\backordering,0,\toPod,\toPickOne,\toPickTwo,\pickerQueueOne,\pickerQueueTwo,\toStorageOne,\toStorageTwo,\replQueueOne, \replQueueTwo,\toReplOneToStorage, \toReplTwoToStorage\right):\\
& \phantomeq\phantom{\cup\big\{\:}\backordering\in\mathbb{N},\:n_{j}\in\left\{ 0,\ldots,\robotnumb\right\} \ \forall j\in\nodeSet,\ \sum_{j\in\nodeSet}n_{j}=\robotnumb\big\}.
\end{align*}

%-------------

\subsection{Calculation of the order turnover time} \label{subsec:turnover}

The turnover time  of a task is measured from the time the task is
received to the time the picking is completed. Therefore, the turnover time can be split into two main parts:
\begin{enumerate}
    \item the waiting time in the external queue and
    \item the processing time in the inner network.
\end{enumerate} 

Note that an order can contain several items stored in different pods. In other words, all the items in an order may need to be completed by several robot tasks. When calculating the turnover time, we ignore the waiting time of an order between the first picked item and its completion. There are two reasons for this: first, the waiting time depends on the efficiency of other algorithms, such as the order of the pods (see \cite{Boysen.2017}); second, we assume that the shorter the waiting time of each item in an order, the shorter the waiting time of the order.

\begin{figure}[h]
	\includegraphics[width=1\textwidth]{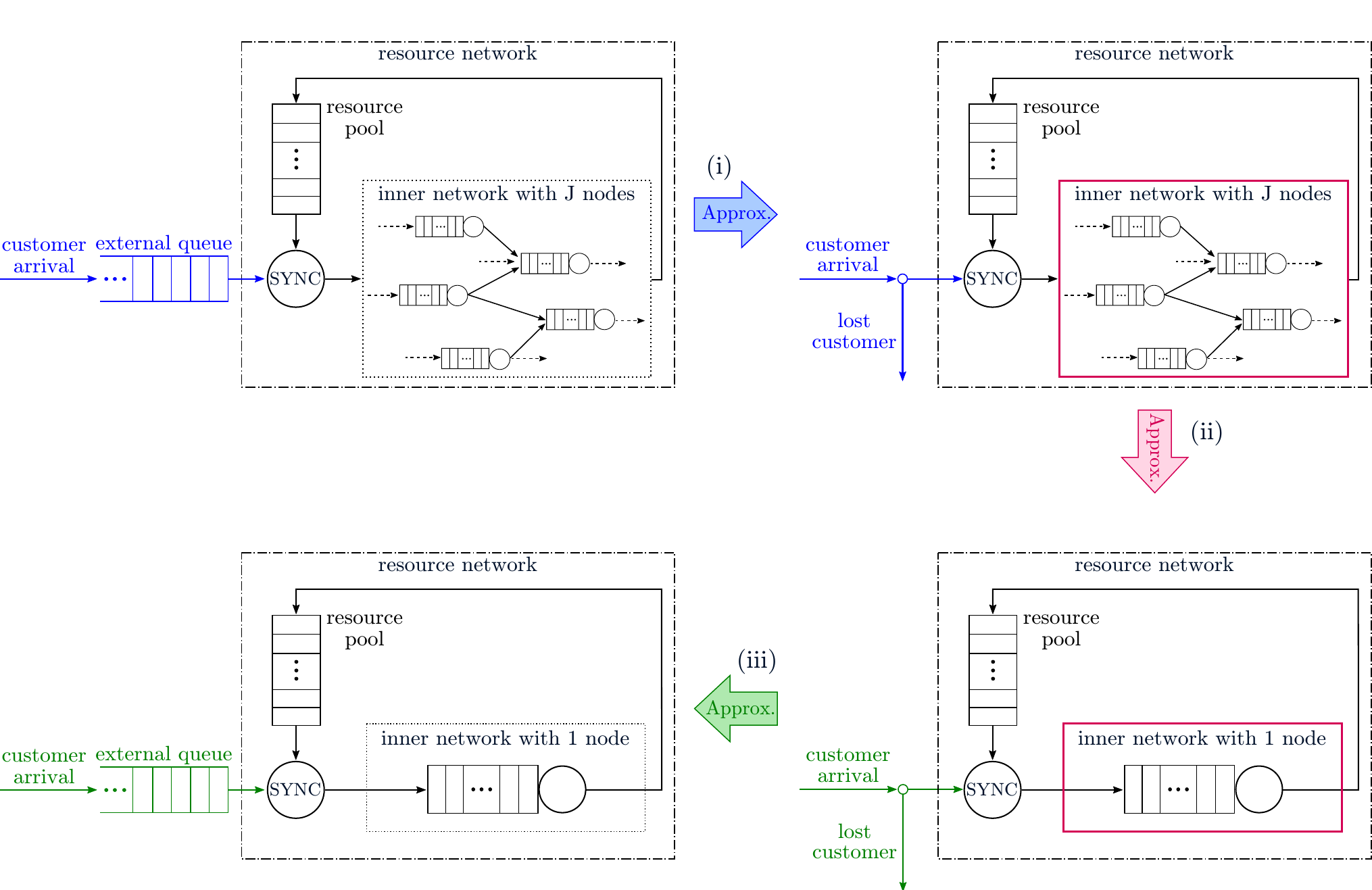}

 %--------------
 
	\caption{Overview of the solution approach developed in \cite{otten2019numRobot}. We use the same colour for parts that we change in a single approximation step. (This figure is a slight modification of~\cite[Fig.~1, p.~605]{otten2019numRobot}.}
	\label{fig:overview-models}
\end{figure}

Because of the large state space, it is impractical to solve the RMFSs exactly using matrix-analytic methods. Even for 10 robots and 11 inner nodes, we need a special matrix with $\binom{J+N-1}{N}^2 =34,134,779,536$ entries.
Therefore, we use the approximation method for SOQN developed by~\cite{otten2019numRobot} to estimate the main performance metrics. An overview of this solution approach is visualised in Figure~\ref{fig:overview-models}. 

%-----
To compute the processing time in the inner network, in (i) in Figure \ref{fig:overview-models} the system is modified such that new arrivals are lost if the resource pool is empty.
Since closed-form expressions for the steady-state distribution
in product form are available for this modification, this approximation can be used to calculate the processing time of the tasks by using mean-value analysis (MVA).

To compute the waiting time in the external queue, in (ii) in Figure \ref{fig:overview-models} the complexity of the
modified SOQN  is reduced by using Norton’s theorem, and then in (iii) in Figure \ref{fig:overview-models}   the external
queue is reinvented.
 Due to its closed-form expressions for the steady-state distribution, the average waiting time in the external queue can be computed using Little's law. 
 
 The details of the approximations and the proofs can be found in \cite{otten2019numRobot}. The quality of the approximations has been confirmed by the simulation results.

\begin{table}[h]  
    \caption{Overview of the nodes in the networks of RMFS.}
		\begin{tabular}{ccccc}
			\hline 
			\multirow{2}{*}{\textbf{Node}} & \textbf{Service} & \textbf{Random} & \multirow{2}{*}{\textbf{State}} & \textbf{Description}\tabularnewline
			& \textbf{intensity} & \textbf{variable} &  & \textbf{(number of robots at time $t$)}\tabularnewline
			\hline 
			\multirow{2}{*}{$\toPodState$} & \multirow{2}{*}{$\transToPod\cdot\phiToPod(\toPod)$} & \multirow{2}{*}{$\procToPod(t)$} & \multirow{2}{*}{$\toPod$} & \multirow{2}{*}{moving in the storage area  to a pod} \tabularnewline
			&  &  &  & \tabularnewline
			\hline 
			\multirow{2}{*}{$\toPickOneState$} & \multirow{2}{*}{$\transToPickerOne\cdot\phiToPickOne(\toPickOne)$} & \multirow{2}{*}{$\procToPickOne(t)$} & \multirow{2}{*}{$\toPickOne$} & moving a pod  from the storage area\tabularnewline
			&  &  &  & to picking station $1$\tabularnewline
			\hline 
			\multirow{2}{*}{$\toPickTwoState$} & \multirow{2}{*}{$\transToPickerTwo\cdot\phiToPickTwo(\toPickTwo)$} & \multirow{2}{*}{$\procToPickTwo(t)$} & \multirow{2}{*}{$\toPickTwo$} & moving a pod  from the storage area\tabularnewline
			&  &  &  & to picking station $2$\tabularnewline
			\hline 
			\multirow{2}{*}{$\pickerQueueOneState$} & \multirow{2}{*}{$\pickingOne$} & \multirow{2}{*}{$\procPickerQueueOne(t)$} & \multirow{2}{*}{$\pickerQueueOne$} & \multirow{2}{*}{in the queue of picking station $1$}\tabularnewline
			&  &  &  & \tabularnewline
			\hline 
			\multirow{2}{*}{$\pickerQueueTwoState$} & \multirow{2}{*}{$\pickingTwo$} & \multirow{2}{*}{$\procPickerQueueTwo(t)$} & \multirow{2}{*}{$\pickerQueueTwo$} & \multirow{2}{*}{in the queue of picking station $2$}\tabularnewline
			&  &  &  & \tabularnewline
			\hline 
			\multirow{3}{*}{$\toStorageOneState$} & \multirow{3}{*}{$\transToStorageOne\cdot\phiToStorageOne(\toStorageOne)$} & \multirow{3}{*}{$\procToStorageOne(t)$} & \multirow{3}{*}{$\toStorageOne$} & moving a pod  from \tabularnewline
			&  &  &  &picking station $1$ to the storage area\tabularnewline
			&  &  &  & and entering node $0$\tabularnewline
			\hline 
			\multirow{3}{*}{$\toStorageTwoState$} & \multirow{3}{*}{$\transToStorageTwo\cdot\phiToStorageTwo(\toStorageTwo)$} & \multirow{3}{*}{$\procToStorageTwo(t)$} & \multirow{3}{*}{$\toStorageTwo$} & moving a pod from \tabularnewline
			&  &  &  & picking station $2$ to the storage area\tabularnewline
			&  &  &  & and entering node $0$\tabularnewline
			\hline 
			\multirow{3}{*}{$\toReplOneStateNew$} & \multirow{3}{*}{$\transToReplOneNew\cdot\phiToReplOneNew(\toReplOne)$} & \multirow{3}{*}{$\procToReplOneNew(t)$} & \multirow{3}{*}{$\toReplOneNew$} & moving a pod \tabularnewline
			&  &  &  & from picking station $1$\tabularnewline
			&  &  &  & to the replenishment station $1$\tabularnewline
			\hline 
			\multirow{3}{*}{$\toReplTwoStateNew$} & \multirow{3}{*}{$\transToReplTwoNew\cdot\phiToReplTwoNew(\toReplTwoNew)$} & \multirow{3}{*}{$\procToReplTwoNew(t)$} & \multirow{3}{*}{$\toReplTwoNew$} & moving a pod\tabularnewline
			&  &  &  & from picking station $2$\tabularnewline
			&  &  &  & to the replenishment station $2$\tabularnewline
			\hline 
			\multirow{2}{*}{$\replQueueOneState$} & \multirow{2}{*}{$\replOne$} & \multirow{2}{*}{$\procReplOne(t)$} & \multirow{2}{*}{$\replQueueOne$} 
			& in the queue \tabularnewline
			&  &  &  & of the replenishment station $1$  \tabularnewline
			\hline 
            \multirow{2}{*}{$\replQueueTwoState$} & \multirow{2}{*}{$\replTwo$} & \multirow{2}{*}{$\procReplTwo(t)$} & \multirow{2}{*}{$\replQueueTwo$} 
			& in the queue \tabularnewline
			&  &  &  & of the replenishment station $2$  \tabularnewline
			\hline 
			\multirow{4}{*}{$\toReplOneToStorageState$} & \multirow{4}{*}{$\transReplOneToStorage\cdot\phiToStorageOne(\toReplOneToStorage)$} & \multirow{4}{*}{$\procReplOneToStorage(t)$} & \multirow{4}{*}{$\toReplOneToStorage$} & moving a pod \tabularnewline
			&  &  &  & from the replenishment station $1$\tabularnewline
			&  &  &  & to the storage area \tabularnewline
			&  &  &  & and entering node $0$\tabularnewline
			\hline 
   			\multirow{4}{*}{$\toReplTwoToStorageState$} & \multirow{4}{*}{$\transReplTwoToStorage\cdot\phiToStorageTwo(\toReplTwoToStorage)$} & \multirow{4}{*}{$\procReplTwoToStorage(t)$} & \multirow{4}{*}{$\toReplTwoToStorage$} & moving a pod \tabularnewline
			&  &  &  & from the replenishment station $2$\tabularnewline
			&  &  &  & to the storage area \tabularnewline
			&  &  &  & and entering node $0$\tabularnewline
			\hline 
	\end{tabular}	
     \label{tab:model-nodes-table}
\end{table}

 \section{Experiment and results} \label{sec:results}
In this section, the parameters used in our experiment are first listed in Subsection~\ref{subsec:parameter} and the computational results of our experiment are shown in Subsection~\ref{subsec:results}.
 \subsection{Parameter settings}\label{subsec:parameter}
 
In our experiments we take parameters from \cite[Section 5.4]{otten2019numRobot}. \label{page:Experiment}
The maximal number of pods is $\robotnumb^{\max}=550$,
arrival rate of tasks is $468\:\frac{\text{tasks}}{\text{h}}=0.13\:\frac{\text{tasks}}{\text{s}}$,
arrival rates are given in {[}order/hour{]}. Because every order generates one task, we use {[}task/hour{]} directly.
Average travel time at node $\toPodState$: $\transToPod^{-1}=\SI{18.4}{\second}$,
%average travel time 
at node $\toPickOneState$: $\transToPickerOne^{-1}=\SI{34.5}{\second}$,
%average travel time 
at  node $\toPickTwoState$: $\transToPickerTwo^{-1}=\SI{34.5}{\second}$,
%average travel time 
at node $\toStorageOneState$: $\transToStorageOne^{-1}=\SI{34.5}{\second}$,
%average travel time 
at node $\toStorageTwoState$: $\transToStorageTwo^{-1}=\SI{34.5}{\second}$,
%average travel time 
at node $\toReplTwoStateNew$: $\transToReplTwoNew^{-1}=\SI{34.5}{\second}$,
%average travel time 
at node $\toReplOneStateNew$: $\transToReplOneNew^{-1}=\SI{34.5}{\second}$,
%and average travel time 
at node $\toReplToStorageState$: $\transReplToStorage^{-1}=\SI{34.5}{\second}$.
Average picking time of picking stations $1$ and $2$: $\pickingOne^{-1}=\pickingTwo^{-1}=\SI{10}{\second}$,
%average pick time 
average replenishment time of replenishment stations 1 and 2 (node $\replQueueOneState, \replQueueTwoState$): $\replOne^{-1}=\replTwo^{-1}=\SI{30}{\second}$. The probability of visiting picking stations 1 and 2: $\transprobPickOne=\transprobPickTwo=0.5$ and the probability of visiting replenishment stations 1 and 2: $\transprobReplOne=\transprobReplTwo=0.2$.
We assume that moving robots do not interfere.
Hence, our processor-sharing queues are infinite server
queues, i.e.~$\phi_{j}(n_{j})=n_{j}$ for all $j\in\nodeSet\setminus\left\{ \pickerQueueOneState,\pickerQueueTwoState,\replQueueOneState, \replQueueTwoState\right\}$.
 \subsection{Results}\label{subsec:results}

\begin{figure}
    \centering
    \includegraphics[width=0.75\textwidth]{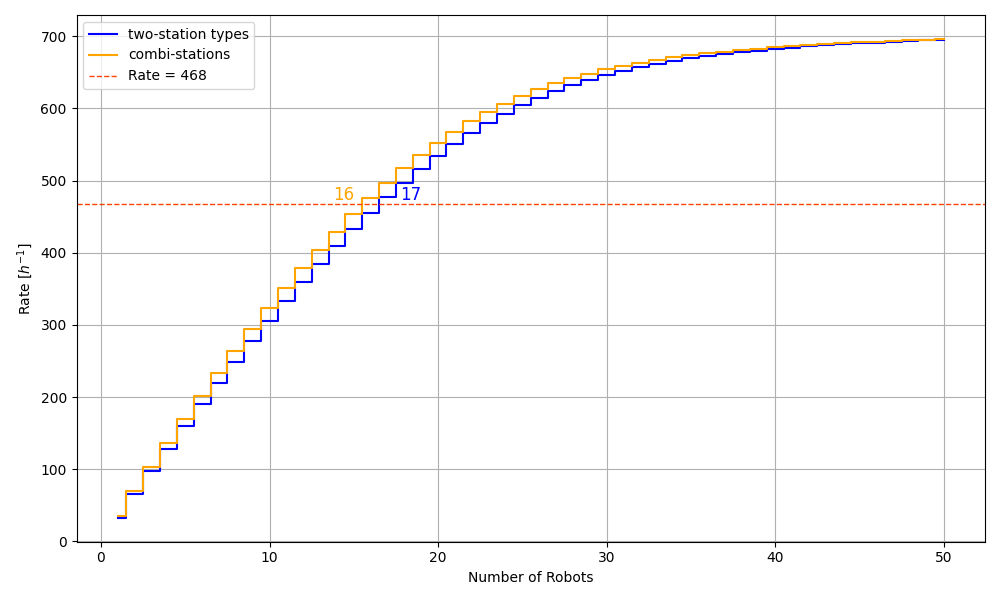}
    \caption{Maximal arrival rates $\arrival$ for given numbers of robots to keep the system stable.}
    \label{fig:stable-arrivals}
\end{figure}

\begin{figure}
    \centering
    \includegraphics[width=0.75\textwidth]{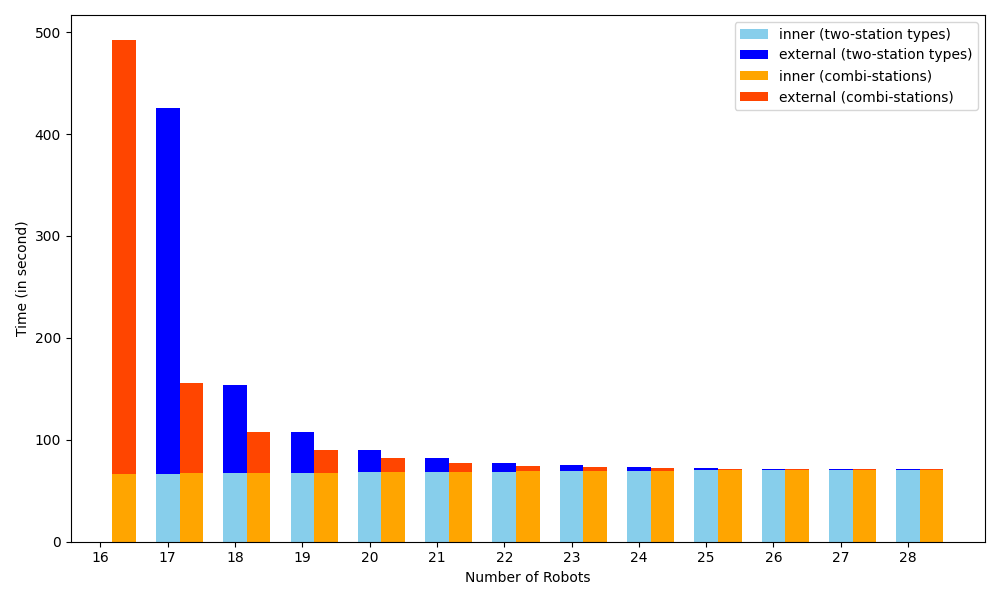}
    \caption{Average turnover time $\TOtask(\arrivalLS,\poolSize)$ of a task (in seconds) for the given numbers of robots. Note that the system of two-station types is not stable with 16 robots, so there is no calculation for it.}
    \label{fig:waiting-times}
\end{figure}

Same as in \cite{otten2019numRobot}, we implemented our algorithm in R and used the \emph{queueing} package,
see \cite{queueingpackage}. Our implementation runs on a PC with an Intel Core i7-7700K CPU @4.20GHz and 32GB RAM running Microsoft Windows 10 in 64-bit mode. We have plotted important system
parameters in Figures \ref{fig:stable-arrivals} and
\ref{fig:waiting-times} respectively. For better readability, we have plotted data
for a limited number of robots (until stabilising of curves).

\prettyref{fig:stable-arrivals} shows maximal arrival rates $\arrival$
for a given number of $\robotnumb$ robots  
to keep the system stable, i.e. the system can ensure that tasks are processed in a timely manner without causing overload or excessive queueing. When using two-station types (each type has two stations), the minimum number of robots for the system to be stable is 17, while the number is reduced to 16 when using two combi-stations. In particular, with more than 40 robots, it is not possible to significantly increase the arrival rate with additional robots. It is worth noting that the RMFS example shown in \cite[Section 5]{otten2019numRobot} is slightly different, as they consider only one replenishment station (whereas we consider two). However, it is interesting to note that with the addition of one replenishment station, we need one less robot. 

\prettyref{fig:waiting-times} shows the average turnover times, i.e. from the arrival of an order (= task) at the system to its completion at a picking station. Note that the time the robot spends travelling to and from the picking station is not counted. For this reason, the time spent in the internal network is identical for both systems. However, the effect of the saving in travel time from picking to replenishment can be seen in the extreme reduction in waiting time in the external network, as the robots are available in the resource pool more quickly. If we invest in 17 robots for the example given, both systems are stable, but the turnover time of the two-station types system is extremely high. It is worth noting that with the same number of robots, the turnover time of the combi-stations system decreases dramatically (by about 64\%). If the number of robots is increased, e.g.~to 18, the average order turnover time can still be reduced by about 30\% in the combi-station system.\\ 

 \section{Conclusion and recommendation} \label{sec:concl}
 In a robotic mobile fulfilment system, the movement through the warehouse area is performed by mobile robots to relieve the traveling loads of human workers. The human workers can only work at workstations, either picking or replenishment stations (depending on the tasks). Due to this high level of automation in warehouses, we have seen several practical examples where the placement of different types of stations can be more flexible. The traditional way is to place the replenishment stations close to the inbound (pallet) gate, while the picking stations are closer to the outbound (parcel) gate. In this paper we introduce a new type of workstation, the combi-station. In such a station, both picking and replenishment tasks can be performed by two different human workers. We can think of a combi-station as capitalising on the dual functionality of picking and replenishment within a single workstation. 

 In order to analyse the performance of using this new type of workstations, we model the two-station types system and the combi-station system (both systems have two picking stations/parts and two replenishment stations/parts) as semi-open queueing networks with backordering. Furthermore, we apply the approximation methods proposed by \cite{otten2019numRobot} to calculate the order turnover time, namely the waiting time in the external queue and the processing time in the inner network. 

 Based on the experiment of the example shown in \prettyref{fig:rmfs}, we can conclude that by replacing workstations with combi-stations, the number of robots needed to reach the stable state is reduced by one. Furthermore, if we keep the same minimum number of robots of the two-station type system (namely 17 robots), the average order turnover time can be reduced by about 64\% in the combi-station system.

 The recommendation for those warehousing companies designing new (automated) warehouses or owning a robotic mobile fulfilment system is to change the traditional way of placing picking and replenishment stations separately, namely integrating combi-stations to capitalise on the dual functionality of picking and replenishment within a single workstation. Such a change in layout can actually speed up order processing time within warehouses. This is exactly what customers expect to get their parcels faster.  

 To investigate the long-term benefits and potential scalability of combi-stations in larger RMFS setups, further studies are recommended.
\bibliographystyle{apalike}
 \bibliography{biblio}
\end{document}